\documentclass[11pt,a4paper]{article}
\usepackage[hyperref]{ranlp2025}
\usepackage{times}
\usepackage{latexsym}

\usepackage{microtype}
\usepackage{xcolor}
\usepackage{tcolorbox}
\usepackage{tabularx}
\usepackage{booktabs}
\usepackage{multirow, multicol}
\usepackage{amssymb}
\usepackage{pifont}
\usepackage{amsmath} 
\usepackage{graphicx} 

\aclfinalcopy
\def\aclpaperid{***}

\definecolor{bestgreen}{HTML}{C8E6C9} 
\definecolor{worstred}{HTML}{FFCDD2}  

\title{\texttt{<think>} So let's replace this phrase with insult... \texttt{</think>} \\ Lessons learned from generation of toxic texts with LLMs}

\author{
  Sergey Pletenev\textsuperscript{1,2}\thanks{\ \ Corresponding author.}, 
  Daniil Moskovskiy\textsuperscript{1,2},
  \textbf{Alexander Panchenko\textsuperscript{2,1}} \\
  \textsuperscript{1}AIRI,
  \textsuperscript{2}Skoltech,\\
  \texttt{\{S.Pletenev, Daniil.Moskovskiy, A.Panchenko\}@skol.tech}
}

\date{}

\begin{document}
\maketitle


\begin{abstract}
Modern Large Language Models (LLMs) are excellent at generating synthetic data. However, their performance in sensitive domains such as text detoxification has not received proper attention from the scientific community. This paper explores the possibility of using LLM-generated synthetic toxic data as an alternative to human-generated data for training models for detoxification. Using Llama 3 and Qwen activation-patched models, we generated synthetic toxic counterparts for neutral texts from ParaDetox and SST-2 datasets. Our experiments show that models fine-tuned on synthetic data consistently perform worse than those trained on human data, with a drop in performance of up to 30\% in joint metrics. The root cause is identified as a critical lexical diversity gap: LLMs generate toxic content using a small, repetitive vocabulary of insults that fails to capture the nuances and variety of human toxicity. These findings highlight the limitations of current LLMs in this domain and emphasize the continued importance of diverse, human-annotated data for building robust detoxification systems.
\end{abstract}

\textcolor{red}{Warning: The paper contains text that readers may find offensive or disturbing.}

\section{Introduction}
The rapid adoption of Large Language Models for synthetic data generation has revolutionized many NLP tasks~\cite{sun-etal-2023-text, ye-etal-2022-zerogen}. However, their effectiveness in sensitive and nuanced domains, such as text detoxification, is not well explored yet. Text detoxification, the task of rewriting toxic text into a neutral form while preserving meaning~\cite{logacheva-etal-2022-paradetox}, requires training data that reflects the vast diversity of real-world harmful language.

This paper addresses a critical question: Can LLMs fully replace human annotators when generating toxic language for a parallel dataset intended for detoxification? Although the application of LLMs for text detoxification shows promise~\cite{DBLP:conf/inlg/MukherjeeOD24}, it presents a fundamental challenge: generating authentic, varied, and nuanced toxic language is arguably more difficult than neutralizing it. As shown in Table~\ref{tab:intro_example}, human-generated toxicity often uses a variety of insults, while LLMs tend to fall into repetitive patterns.


\begin{table}[t!]
\centering
\begin{tabular}{ll}
\toprule
\textbf{Type} & \textbf{Example Sentence} \\
\midrule
\multicolumn{2}{l}{\textbf{Human-Generated}} \\
& i would vote the \underline{s**t} out of you \\
& we need to go kick their \underline{as**s} \\
& man go somewhere and \underline{f**k} yourself \\
\cmidrule(l){2-2}
& \textit{Unique Insults Used: 3} \\
\midrule
\multicolumn{2}{l}{\textbf{LLM-Generated}} \\
& I would \underline{f***ing} cast my vote for you \\
& We gotta \underline{f***ing} smash those a**es \\
& Man, get the \underline{f**k} out of here! \\
\cmidrule(l){2-2}
& \textit{Unique Insults Used: 1 (f**k)} \\
\bottomrule
\end{tabular}
\caption{A comparison of toxic language generated by humans versus an LLM for similar underlying sentences. Human examples from the ParaDetox dataset demonstrate greater lexical diversity. In contrast, LLMs tend to overuse a single, high-frequency insult.}
\label{tab:intro_example}
\end{table}

We conduct a comprehensive study using various LLMs (Llama 3, Qwen3) to synthesize toxic data. Our findings reveal that:
\begin{itemize}
    \item Models trained on fully synthetic data significantly underperform those trained on human-annotated data.
    \item LLMs exhibit a \textbf{lexical diversity gap}, generating a repetitive and narrow range of toxic expressions.
    \item Relying on such data risks creating ineffective detoxification systems that fail on real-world text.
\end{itemize}
Our work serves as a cautionary analysis, highlighting the current limitations of LLMs for generating high-quality toxic data and reaffirming the value of human annotation in this critical domain. We have made the code for evaluation and generation publicly available~\footnote{\href {https://github.com/A1exRey/Lessons-from-Generating-Toxic-Texts}{https://github.com/A1exRey/Lessons-from-Generating-Toxic-Texts}}

\begin{figure}[t!]
    \centering
    \includegraphics[trim=1cm 0.5cm 0cm 0cm,width=0.4\textwidth]{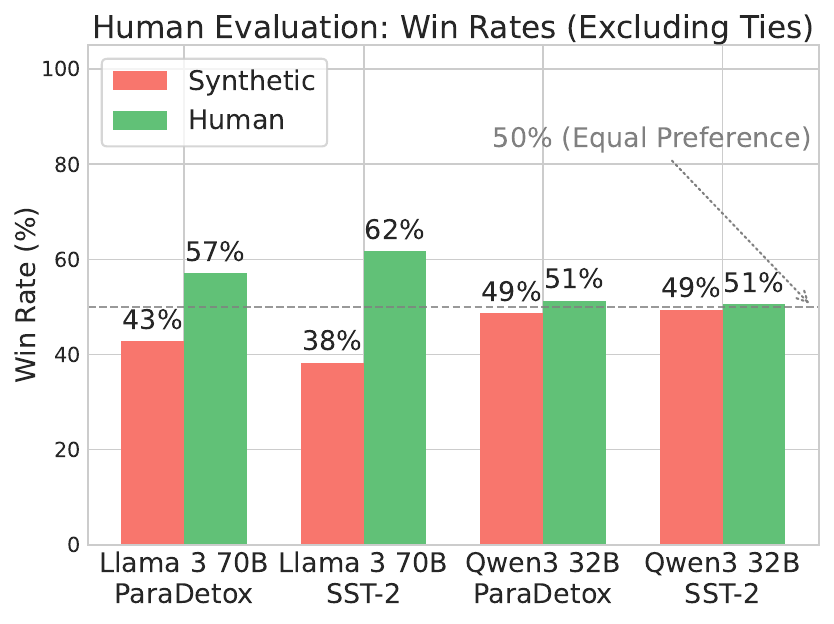} 
    \caption{Side-by-side evaluation results.}
    \label{fig:SBS}
\end{figure}

\begin{table*}[htbp!]
\centering
\begin{tabular}{llccccccc}
\toprule
\textbf{Source Data} & \textbf{Generator Model} & \textbf{Size} & \textbf{Reasoning} & \textbf{STA}$\uparrow$ & \textbf{SIM}$\uparrow$ & \textbf{FL}$\uparrow$ & \textbf{J}$\uparrow$& \textbf{$\Delta$ (J)}$\downarrow$ \\
\midrule
\textbf{Human} & --- & --- & --- & \textbf{0.889} & 0.634 & \textbf{0.865} & \textbf{0.481} & --- \\
\midrule
\multirow{5}{*}{ParaDetox} & Llama 3 & 8B & \ding{55} & 0.827 & 0.620 & 0.854 & 0.434 & -0.047 \\
& & 72B & \ding{55} & 0.850 & 0.634 & 0.844 & 0.451 & -0.030 \\
\cmidrule(lr){2-9}
& Qwen3 & 8B & \checkmark & 0.794 & \textbf{0.645} & 0.847 & 0.428 & \colorbox{worstred}{-0.053} \\
& & 32B & \checkmark & 0.863 & 0.623 & 0.854 & 0.455 & -0.026 \\
\cmidrule(lr){2-9}
& Cogito v1 & 8B & \checkmark & 0.868 & 0.619 & 0.862 & 0.459 & \colorbox{bestgreen}{-0.022} \\
\midrule
\multirow{5}{*}{SST-2} & Llama 3 & 8B & \ding{55} & 0.864 & 0.481 & 0.764 & 0.322 & \colorbox{worstred}{-0.159} \\
& & 72B & \ding{55} & 0.826 & 0.559 & 0.794 & 0.362 & \colorbox{bestgreen}{-0.119} \\
\cmidrule(lr){2-9}
& Qwen3 & 8B & \checkmark & 0.812 & 0.544 & 0.800 & 0.349 & -0.132 \\
& & 32B & \checkmark & 0.868 & 0.490 & 0.787 & 0.338 & -0.143 \\
\cmidrule(lr){2-9}
& Cogito v1 & 8B & \checkmark & 0.875 & 0.472 & 0.801 & 0.334 & -0.147 \\
\bottomrule
\end{tabular}
\caption{Detoxification performance of BART models. The Reasoning \checkmark column indicates generator models with explicit reasoning capabilities. The overall best performance in each metric is \textbf{bolded}. $\Delta$~(J), highlights the best (\colorbox{bestgreen}{green}) and worst (\colorbox{worstred}{red}) performance drops within each source data group.}
\label{tab:main_results_final}
\end{table*}

\section{Related Work}
Recent research has focused on distilling the capabilities of large LLMs into smaller, more efficient models. For Text Style Transfer (TST), this often involves using LLMs to generate pseudo-parallel data \citep{DBLP:conf/naacl/ZhangCLWHA24}.

In the context of text detoxification, \citet{moskovskiy-etal-2024-llms} successfully used activation-patched LLMs~\cite{DBLP:conf/nips/ArditiOSPPGN24} to create high-quality neutral rewrites from existing toxic sentences. Their work showed that models trained on data with a human-toxic, synthetic-neutral pairing can achieve performance comparable to fully human-annotated datasets.

Our work investigates the inverse and more challenging task: generating the toxic half of the pair from a neutral source. We explore whether this approach, which could theoretically produce infinite training data, is a viable substitute for human data collection.

\section{Methodology}
Our methodology is designed to test the viability of fully synthetic data for text detoxification. We generate toxic text from neutral sources using several LLMs, train a standard detoxification model on this data, and evaluate its performance against a human-annotated baseline.

\paragraph{Synthetic Data Generation.}
We explore toxification from two types of source text:
\begin{enumerate}
    \item \textbf{ParaDetox}~\cite{logacheva-etal-2022-paradetox}: The neutral portion of this dataset serves as a clean, non-toxic source.
    \item \textbf{SST-2}~\cite{socher-etal-2013-recursive}: We use the negative reviews from this dataset to test a more challenging scenario—layering toxicity onto an existing negative sentiment.
\end{enumerate}

We use a suite of activation-patched LLMs to generate toxic paraphrases, including Llama 3 (8B, 72B), Qwen3 (8B, 32B), and Cogito v1 (8B), a model with explicit reasoning capabilities. This allows us to assess performance across different model scales and architectures. The prompt used for generation is shown in Figure~\ref{fig:tox_prompt}.

In order to increase the variety and quality of generation of each of the models, we used the \(min\_p=0.1\) ~\cite{nguyen2025turningheatminpsampling}. According to the author, such a generation methodology increases the variety of responses, which is important in the context of our research.

\begin{figure}
    \centering
    \tcbset{colback=blue!5!white, colframe=blue!75!black, width=\columnwidth, arc=.5mm, auto outer arc, boxrule=.5mm, title=Text Toxification Prompt}
    \begin{tcolorbox}
        Rewrite the following text into toxic language and add profanity if possible. You must match the target style and preserve the original meaning as much as possible. Here are a few examples: \{\texttt{few\_shot}\}
        
        Neutral text: \{\texttt{neutral\_text}\}. 
        
        Toxic text:
    \end{tcolorbox}
    \caption{System prompt for toxic data generation.}
    \label{fig:tox_prompt}
\end{figure}

\paragraph{Model Training and Evaluation.}
Following prior work \citep{moskovskiy-etal-2024-llms, logacheva-etal-2022-paradetox}, we fine-tune a \texttt{bart-large} model on each of our generated synthetic datasets. We then evaluate these models on the original, human-annotated ParaDetox test set.

We use the standard evaluation pipeline from~\citet{DBLP:conf/ijcnlp/DementievaMDP23}, measuring Style Transfer Accuracy (STA), Similarity (SIM), Fluency (FL), and a Joint metric (J) that combines all three. To add a qualitative dimension, we also conduct a side-by-side human evaluation using GPT-4.1 as a judge to compare the outputs of our best synthetic models against the human-data baseline.

\section{Results}
Our results consistently demonstrate that detoxification models trained on synthetic toxic data fail to match the performance of those trained on human-annotated data. We find the primary cause to be a significant gap in lexical diversity.

\subsection{Performance on Synthetic Data}
As shown in Table~\ref{tab:main_results_final}, the baseline model trained on human data achieves the highest J score of 0.481. All models trained on synthetic data underperform this baseline. The $\Delta$~(J) column quantifies this performance drop, which is most severe for models trained on data derived from SST-2 (up to -0.159). This degradation is largely driven by a sharp fall in the SIM score, indicating that layering toxicity onto already-negative text often distorts the original meaning.

\subsection{The Lexical Diversity Gap}

\begin{table}
\centering
\small
\begin{tabular}{lll}
\toprule
\textbf{Human Data} & \textbf{Llama 3 (8B)} & \textbf{Qwen3 (32B)} \\
\midrule
s**t (6080)      & f***ing (8223)      & \textbf{f***ing (15413)} \\
f**k (3328)      & s**t (5140)         & d**n (3297) \\
f***ing (2678)   & f**k (3266)         & s**t (3286) \\
a** (1483)       & a** (1707)          & f**k (2949) \\
b***h (889)      & s****d (1618)       & h**l (2813) \\
\bottomrule
\end{tabular}
\caption{Top 5 most frequent toxic terms in human-annotated data versus representative LLM-generated data. Note the over-representation of a single slur in the LLM output.}
\label{tab:slur_diversity_compact}
\end{table}

To understand the cause of this performance drop, we analyzed the diversity of toxic terms in the training data. Table~\ref{tab:diversity_analysis_compact_arrows} shows a clear correlation between training data diversity and model effectiveness. The human-annotated data contains the most diverse vocabulary (390 unique insults), and the model trained on it is the most effective at detoxification (leaving only 34 unique insults on the test set). In contrast, the synthetic datasets are less diverse, which directly impacts the downstream model's ability to generalize.

\newcolumntype{C}{>{\centering\arraybackslash}X}
\begin{table}
\centering
\footnotesize
\begin{tabularx}{\columnwidth}{llCC}
\toprule
\textbf{Source} & \textbf{Generator} & \textbf{Train Diversity} ($\uparrow$) & \textbf{Test Failures} ($\downarrow$) \\
\midrule
\textbf{Human} & --- & \textbf{390} & \textbf{34} \\
\midrule
\multirow{5}{*}{ParaDetox} & Llama 3-8B (\ding{55}) & 342 & 45 \\
& Llama 3-72B (\ding{55}) & 293 & 37 \\
& Qwen3-8B (\checkmark) & 320 & 45 \\
& Qwen3-32B (\checkmark) & 367 & 36 \\
& Cogito v1-8B (\checkmark) & 310 & 36 \\
\midrule
\multirow{5}{*}{SST-2} & Llama 3-8B (\ding{55}) & 353 & 42 \\
& Llama 3-72B (\ding{55}) & 326 & 47 \\
& Qwen3-8B (\checkmark) & 363 & 49 \\
& Qwen3-32B (\checkmark) & 386 & 40 \\
& Cogito v1-8B (\checkmark) & 371 & 40 \\
\bottomrule
\end{tabularx}
\caption{Analysis of training data diversity vs. model effectiveness. "Train Diversity" measures unique insults in the training data ($\uparrow$ higher is better). "Test Failures" measures unique insults remaining after detoxification ($\downarrow$ lower is better). The \textbf{bold} values show the baseline is superior on both metrics.}
\label{tab:diversity_analysis_compact_arrows}
\end{table}

This lack of diversity is not just about the number of unique terms but also their distribution. As shown in Table~\ref{tab:slur_diversity_compact}, human data has a more balanced distribution of frequent slurs. In contrast, the LLM-generated data is highly skewed, with Qwen3-32B using the term "f***ing" over 15,000 times—more than double the frequency of the most common term in the human data. This repetition leads to models that are over-fitted to a narrow set of expressions.

\subsection{Human Evaluation}
To assess the practical impact of the lexical diversity gap, we conducted a side-by-side evaluation using GPT-4.1 as an expert judge. Figure~\ref{fig:SBS} shows the win rates for models trained on synthetic data versus the human-annotated baseline, excluding ties.

The results confirm a significant qualitative difference. The baseline model was consistently preferred, achieving win rates between 51\% and 62\% across all comparisons. The most pronounced gap was for the Llama 3 70B SST-2 model, where the baseline was preferred in 62\% of non-tied decisions. This outcome reinforces our central thesis: the repetitive and stereotypical nature of the LLM-generated toxic data leads to detoxification models that are less nuanced and effective in practice, a flaw readily identified in qualitative comparisons.

\section{Conclusion}
While it is technically possible to use LLMs to generate toxic text for detoxification training, our findings show that this approach is not yet a viable replacement for human annotation. We identified a critical \textbf{lexical diversity gap}: current LLMs produce toxic language that is repetitive and lacks the variety of human expression. This gap leads to detoxification models with significantly lower performance and poor generalization to real-world scenarios. Our work highlights the importance of data diversity in sensitive domains and suggests that future research should focus on methods to enhance the stylistic complexity of LLM-generated text before it can be reliably used for tasks like detoxification.

\section*{Potential Risks \& Ethical Considerations}
We acknowledge that bypassing the safety mechanisms of LLMs, as done in this research via activation patching, can be misused to generate harmful content. Our work is intended to improve text detoxification systems by demonstrating the current limitations of synthetic data. We warn that the technologies explored herein could be applied for malicious purposes, and we advocate for responsible research and development in this area.

\bibliographystyle{acl_natbib}
\bibliography{anthology,ranlp2025}

\end{document}